\documentclass[11pt]{article}
\usepackage[margin=1in]{geometry}
\usepackage[utf8]{inputenc}
\usepackage[T1]{fontenc}
\usepackage{textcomp}
\usepackage{amsmath,amssymb,amsthm,amsfonts}
\usepackage{booktabs}
\usepackage{multirow}
\usepackage{graphicx}
\usepackage[section]{placeins}
\graphicspath{{./}}
\usepackage{xcolor}
\usepackage{array}
\usepackage{enumitem}
\usepackage{tabularx}
\newcolumntype{L}{>{\raggedright\arraybackslash}X}
\usepackage{tikz}
\usetikzlibrary{positioning,arrows.meta,shapes.geometric,calc}
\usepackage[hidelinks]{hyperref}
\usepackage[numbers,sort&compress]{natbib}
\usepackage{authblk}
\usepackage{caption}
\newcommand{\claim}{\textsc{Claim}}
\newcommand{\best}[1]{\textbf{#1}}

\title{\vspace{-6mm}\textbf{The Architect, the Adversary, and the Judge:\\
Closed-Loop Generation of Standards-Aligned Assessment Items at Scale}}

\author[1]{Wenhui Chen\thanks{\texttt{mc35092@um.edu.mo}}}
\author[1]{Ziyao Lin\thanks{\texttt{mc35081@um.edu.mo}}}
\author[2]{Jianlin Chen\thanks{\texttt{202330450231@mail.scut.edu.cn}}}
\author[3]{Peiji Long\thanks{\texttt{longpeiji@gmail.com}}}
\author[1]{Chi Man Vong\thanks{Corresponding author. \texttt{cmvong@um.edu.mo}}}
\affil[1]{University of Macau}
\affil[2]{South China University of Technology}
\affil[3]{Independent Researcher}
\date{}

\begin{document}
\maketitle

\begin{abstract}
Large language models promise scalable authoring of K--12 assessment items, but production deployment means satisfying dozens of psychometric, linguistic, and curricular constraints at once, for every grade, standard, and difficulty level.
We present \claim{} (\textbf{C}losed-\textbf{L}oop \textbf{A}ssessment \textbf{I}tem generation with \textbf{M}ined knowledge), a pipeline coupling a two-stage \emph{generate-then-attack} protocol, in which the model drafts an item as a ``curriculum architect'' and then re-enters the same conversation as a \emph{hostile adversarial reviewer}; \emph{bi-directional} few-shot conditioning on accepted \emph{and} rejected items, the latter carrying the evaluator's diagnosis; and a \emph{knowledge dictionary} of 44{,}844 error-correction rules mined from that feedback and retrieved per standard and item type.
Across 43{,}227 scored items over 755 Common Core ELA standards (grades 1--12), three item types, and ten LLMs, the pipeline reaches a 97.8\% expert-evaluator pass rate on a 9{,}074-item production run, to our knowledge the largest reported corpus of standards-aligned items scored by an automated evaluator.
We then ask what that rate certifies. Re-scoring a stratified sample with three judges from other vendors, blind to the deployed verdict, reproduces the format ordering under every judge and recovers a \emph{larger} open-set deficit than the deployed evaluator does; but agreement on the accept/reject binary is weak at production prevalence ($\kappa{\approx}0.13$), and the judges agree with each other no better. The level is therefore judge-relative, and with no student-response data our quality evidence is evaluator-judged throughout \citep{goel2025aiexams}.
The corpus also exposes a robust asymmetry: multiple-choice and multiple-select generation saturate at ${\geq}98\%$ for both frontier models under a dozen static rules, whereas fill-in-the-blank generation is capability-tiered (82.8--96.7\% across five models under a matched rule set, standards, and judge) and plateaus under prompt-only optimization, with error mass shifting between answer-key \emph{over-inclusion} and \emph{omission} as rules accumulate.
We analyze this as an \emph{open-set boundary determination} problem that autoregressive decoders are structurally ill-equipped to solve, and read the round-by-round error shift as suggestive of a precision--recall trade-off rather than as a demonstrated frontier.
Distilling 40K+ expert verdicts into an open-weights judge reproduces the asymmetry \emph{inside} the evaluator: the student acquires the verdict form but not its discrimination, fail-recall rising 8${\to}$63\% while F1 saturates at 0.25 and pass/fail agreement never reaches the teacher's own consistency band.
\end{abstract}

\section{Introduction}
\label{sec:intro}

Every adaptive-learning platform needs a continuous supply of assessment items that are aligned to a curriculum standard, calibrated to a difficulty level, and free of the item-writing flaws that psychometricians have catalogued for decades \citep{haladyna1989taxonomy,haladyna2002review}.
Hand-authoring at this specification is slow and expensive; template-based automatic item generation \citep{gierl2012aig,gierl2013book} scales but confines items to narrow structural molds.
Large language models (LLMs) can author fluent, curriculum-referenced items \citep{elkins2023useful,sauberli2024rc}, yet two obstacles separate a fluent item from a \emph{deployable} one.
First, deployability is conjunctive: an item must simultaneously satisfy standard alignment, difficulty calibration, distractor diagnosticity, answer-key exactness, factual accuracy, and cultural neutrality---and LLMs are known to degrade as constraints accumulate \citep{jiang2024followbench,wen2024complexbench,jaroslawicz2025ifscale}.
Second, quality is defined externally: whether a generated answer key is \emph{complete} and \emph{exclusive} is adjudicated against the semantics of the target language and curriculum, not against the model's own output distribution.

\begin{figure}[t]
\centering
\resizebox{\linewidth}{!}{%
\begin{tikzpicture}[
  font=\small,
  node distance=4mm and 7mm,
  box/.style={draw=black!60, rounded corners=1.5pt, align=center, inner sep=4pt, minimum height=8.5mm},
  store/.style={box, fill=blue!6},
  gen/.style={box, fill=orange!8},
  judge/.style={box, fill=green!8},
  arr/.style={-{Stealth[length=2mm]}, thick, black!70},
  feed/.style={-{Stealth[length=2mm]}, thick, red!60!black, dashed}
]
\node[store] (curr) {CCSS curriculum\\755 standards, G1--12};
\node[store, below=of curr] (kd) {Knowledge dictionary\\44{,}844 mined rules\\keyed by (standard, type)};
\node[store, below=of kd] (ex) {Exemplar pools\\23{,}530 accepted / 2{,}992 rejected\\(with evaluator diagnoses)};
\node[gen, right=9mm of kd, text width=34mm] (p1) {\textbf{Stage 1: Architect}\\4-phase protocol\\\textsc{analyze} $\to$ \textsc{blueprint}\\$\to$ \textsc{draft} $\to$ \textsc{validate}\\(12-gate checklist)};
\node[gen, right=7mm of p1, text width=33mm] (p2) {\textbf{Stage 2: Adversary}\\same conversation,\\hostile reviewer\\5 surgical attacks\\$\to$ corrected JSON};
\node[judge, right=7mm of p2, text width=29mm] (ib) {\textbf{Judge}\\expert evaluator v2.3.3\\multi-dimension score\\pass iff $s > 0.85$};
\node[box, above=4mm of ib, fill=black!4] (bank) {Item bank\\(pass)};
\draw[arr] (curr.east) -| ($(p1.west)+(-3mm,8mm)$) -- ($(p1.west)+(0,8mm)$);
\draw[arr] (kd) -- (p1);
\draw[arr] (ex.east) -| ($(p1.west)+(-3mm,-8mm)$) -- ($(p1.west)+(0,-8mm)$);
\draw[arr] (p1) -- (p2);
\draw[arr] (p2) -- (ib);
\draw[arr] (ib) -- (bank);
\coordinate (fb) at ($(ex.south)+(0,-6mm)$);
\coordinate (lft) at ($(ex.west)+(-5mm,0)$);
\draw[feed] (ib.south) -- (ib.south |- fb) -- node[above, midway, font=\footnotesize\itshape]{fail: mine rules \& exemplars from evaluator feedback} (fb) -- (ex.south);
\draw[feed] (fb) -| (lft |- kd.west) -- (kd.west);
\end{tikzpicture}}
\caption{The \claim{} closed loop. A blueprint contract (grade, standard, type, difficulty) is compiled into a prompt together with retrieved knowledge rules and contrastive exemplars. Stage 1 drafts under a four-phase protocol; Stage 2 re-enters the same conversation as a hostile reviewer. Every evaluator verdict feeds back into the exemplar pools and the knowledge dictionary used by future generations.}
\label{fig:pipeline}
\end{figure}
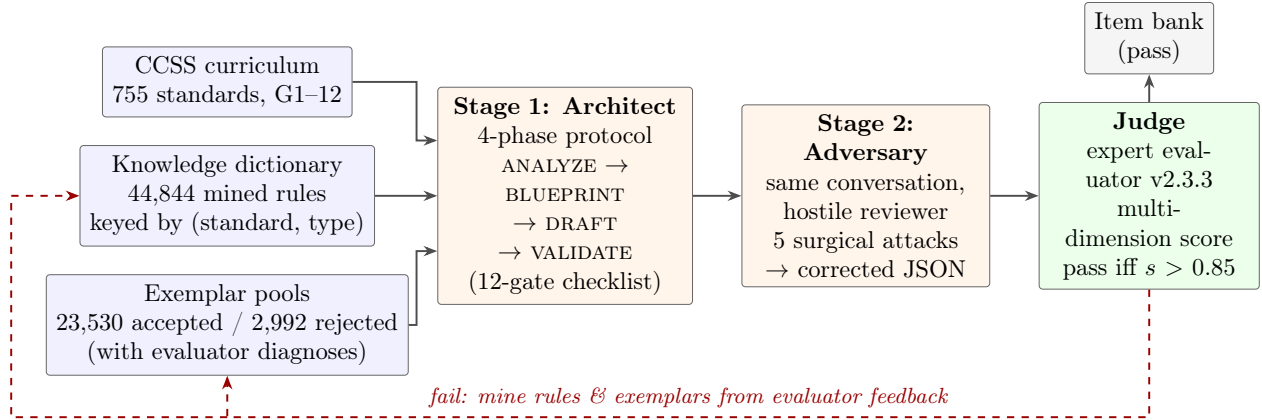

This paper reports what it took to push an LLM item-generation system to production quality across an entire K--12 English Language Arts (ELA) curriculum, and what systematically resisted that push.
Our pipeline, \claim{} (Figure~\ref{fig:pipeline}), closes the loop between a generator and a multi-dimensional expert evaluator: every evaluator verdict is recycled, either as a contrastive few-shot exemplar (accepted \emph{and} rejected items, the latter with the evaluator's diagnosis attached) or as a mined \emph{knowledge rule}---an error-correction pair indexed by curriculum standard and item type and re-injected into future prompts.
Generation itself is two-staged: a ``Chief Curriculum Architect'' drafts the item under a four-phase protocol with an internal quality-gate checklist, then---in the same conversational context, with the draft and its reasoning trace visible---the model is re-instantiated as a \emph{hostile adversarial reviewer} whose only mission is to break the draft: prune leaked synonyms, test every answer-key entry by substitution, audit each factual claim in the explanation.

Run at scale, the pipeline produced 65{,}735 item instances over three weeks of iterative development; 43{,}227 scored items (47{,}278 verdicts) across ten LLMs form the empirical basis of this paper.
The headline result is positive: the final production run---9{,}074 items covering 755 Common Core standards $\times$ three item types $\times$ three difficulty levels over grades 1--12---passes the expert evaluator at 97.8\%, with every grade$\times$type cell confined to a 4.3-point band (95.3--99.6\%) across the entire curriculum.

The more interesting result is negative.
Multiple-choice (MCQ) and multiple-select (MSQ) items proved \emph{easy}: a single set of ${\sim}12$ static format rules held both frontier models at ${\geq}98\%$ with no iterative tuning, and held every grade and difficulty of the production run at ${\geq}98.0\%$ on MCQ and ${\geq}96.3\%$ on MSQ.
Fill-in-the-blank (cloze) items did not.
Under identical scaffolding they required four rounds of rule accretion (15${\to}$42 rules), remained 2.8--16.8 points below the same model's MCQ rate for every model measured on both, separated models into sharp capability tiers (82.8--96.7\% in the matched comparison of Table~\ref{tab:g12}), and exhibited a signature failure dynamic: rules that suppressed answer-key \emph{over-inclusion} inflated answer-key \emph{omission}, and vice versa, which we read as movement \emph{along} a precision--recall frontier rather than beyond it.
We formalize the asymmetry as \emph{closed-set self-referential} generation (the MCQ answer space is constructed by the model itself, so the judge checks internal consistency) versus \emph{open-set boundary determination} (the cloze answer key must reproduce a boundary defined by the language and curriculum, so the judge checks external correctness), and connect the observed error modes to known structural properties of autoregressive decoders: sequential set enumeration without lookahead \citep{vinyals2016order,welleck2020consistency,hou2025enumeration}, predictive rather than discriminative representations \citep{west2024paradox,jacobs2024cloze}, and uncalibrated set-membership thresholds \citep{kadavath2022know,xiong2024confidence}.

\paragraph{Contributions.}
\begin{enumerate}[leftmargin=1.2em,itemsep=1pt,topsep=2pt]
\item \textbf{A closed-loop production pipeline} for standards-aligned item generation combining two-stage adversarial self-review, bi-directional contrastive few-shot conditioning, and automatic mining of a 44{,}844-rule knowledge dictionary from evaluator feedback (\S\ref{sec:pipeline}).
\item \textbf{An empirical study at full-curriculum scale}: 43{,}227 scored items (47{,}278 evaluator verdicts) over 755 CCSS ELA standards, grades 1--12, three item types, three difficulty levels, and ten LLMs; the final production run reaches 97.8\% pass (\S\ref{sec:results}). We believe this is the largest reported item corpus scored against curriculum standards, while noting that it carries no student-response evidence \citep{goel2025aiexams}.
\item \textbf{A validity check on the measurement instrument}: three judges from other vendors, blind to the deployed verdict, re-score a stratified 390-item sample. The format asymmetry survives the change of judge, vendor and rubric and is larger outside the deployed evaluator than inside it; agreement on the accept/reject binary is weak at production prevalence, which bounds what any single-judge pass rate certifies (\S\ref{sec:reliability}).
\item \textbf{A characterization of the closed-set/open-set asymmetry}: multi-model evidence that MCQ/MSQ generation is format-saturated while cloze generation is capability-tiered and plateaus under prompt-only optimization, with a mechanistic analysis and an error taxonomy whose reliability we measure rather than assume (\S\ref{sec:results}--\S\ref{sec:analysis}).
\item \textbf{A cross-task corroboration}: distilling the expert evaluator itself into an open-weights student reproduces the generation-over-discrimination asymmetry---the student acquires the verdict \emph{form} but slides along a precision--recall trade-off (fail-recall 8${\to}$63\%, F1 saturating at 0.25) rather than acquiring the accept/reject boundary, isolating the discriminative-representation mechanism (\S\ref{sec:distill}).
\item \textbf{Design implications} for generation-evaluation ecosystems: rubric-based key adjudication over exhaustive key enumeration, externalized boundary verification, and cloze generation as a discriminative stress test that spreads a model tier over 13.9 points where the closed-set formats compress it into 3.9 near ceiling (\S\ref{sec:discussion}).
\end{enumerate}

\section{Related Work}
\label{sec:related}

\paragraph{Automatic item generation.}
Template-based AIG \citep{gierl2012aig,gierl2013book} and early neural approaches \citep{vondavier2018rnn,kurdi2020systematic} established the field's quality bar: items must observe validated item-writing guidelines \citep{haladyna2002review}, and candidate over-generation demands automated ranking \citep{heilman2010good}. Transformer-based AIG reached operational deployment in the Duolingo English Test \citep{attali2022interactive}.
LLM-based item generation has since been assessed for teacher-perceived quality \citep{elkins2023useful}, Bloom-level control \citep{scaria2024bloom}, difficulty control \citep{tomikawa2026difficulty}, and reading-comprehension items \citep{sauberli2024rc,doughty2024comparative}; \citet{wang2024edusurvey} survey the space.
Distractor generation---the classic hard part of MCQ authoring---has its own literature \citep{alhazmi2024survey}, from PLM ranking over cloze corpora \citep{xie2018cloth,chiang2022cdgp} to retrieval-prompted LLMs \citep{bitew2023distractor} and misconception-aware generation \citep{feng2024mathdistractor,fernandez2024divert}. Our knowledge dictionary is conceptually adjacent to DiVERT's learned error representations \citep{fernandez2024divert}, but is mined from \emph{evaluator} feedback rather than student responses, and is retrieved per curriculum standard.
For standards grounding, \citet{lucy2024mathfish} show LLMs struggle to verify alignment to K--12 standards, and \citet{imperial2024standardize} inject standard-derived knowledge artifacts into generation---our per-standard rule retrieval extends this idea with automatically mined, failure-driven artifacts.

\paragraph{LLM evaluators.}
We rely on a commercial multi-dimensional expert evaluator; the paradigm follows LLM-as-judge \citep{zheng2023judging} and rubric-conditioned scoring \citep{liu2023geval,kim2024prometheus}. For MCQ quality specifically, GPT-4 and rule-based detectors recover most item-writing flaws found by human annotators \citep{moore2023assessing,moore2024saquet}; difficulty prediction from text remains hard \citep{yaneva2024bea}. EduBench \citep{xu2025edubench} is the closest public multi-scenario evaluation suite.

\paragraph{Prompt optimization and learning from mistakes.}
Automatic prompt search \citep{zhou2023ape,yang2024opro}, textual-gradient methods \citep{pryzant2023protegi,yuksekgonul2024textgrad}, pipeline compilers \citep{khattab2024dspy}, and adversarial in-context optimization \citep{do2024advicl} all optimize prompts against a scorer.
Our loop differs in that the artifacts extracted from failures are \emph{persistent and structured}: mistake-derived principles \citep{zhang2024leap,sun2024ricp}, deployment-time feedback memories \citep{madaan2022memprompt}, experiential insights \citep{zhao2024expel}, and contrastive negative demonstrations \citep{gao2024contrastive} are the closest precedents; we operationalize all three (rules, memories, negatives) in one production system and report their behavior at scale.

\paragraph{Self-correction.}
Self-refinement helps when feedback is external and specific \citep{madaan2023selfrefine,gou2024critic} and fails when it is intrinsic \citep{huang2024cannot}; principle-conditioned critique \citep{bai2022constitutional} and LLM-vs-LLM red-teaming \citep{perez2022redteam} inform our Stage-2 design, which arms the reviewer with named attacks rather than a generic ``check your work'' instruction.

\paragraph{Set generation and the generative--discriminative gap.}
Autoregressive decoders impose an order on unordered outputs \citep{vinyals2016order,welleck2019nonmonotonic} and lack principled stopping criteria \citep{welleck2020consistency}; recent work documents enumeration deficits directly \citep{hou2025enumeration}.
Calibration is strong on closed answer formats but collapses on ambiguous or boundary cases \citep{kadavath2022know,xiong2024confidence,liu2023ambiguity}; models' token distributions misestimate human cloze acceptability \citep{jacobs2024cloze} and handle fine-grained collocation categories poorly \citep{espinosa2021collocations}.
Generation outstrips discrimination in LLMs \citep{west2024paradox,saunders2022selfcritiquing,jiang2024selfincorrect}.
Our contribution is to show these laboratory phenomena surfacing as the binding production constraint in a commercially consequential task, and to quantify how far prompt-side optimization gets before it stalls.
\citet{chen2026judging} isolate the same split on references whose acceptable set is decidable, so an authored key can be scored exactly against it; the reference here is a curriculum and a language, which is what makes the deficit consequential and also what makes it hard to measure.

\section{Method}
\label{sec:pipeline}

\subsection{Task and evaluation protocol}
\label{sec:task}
Each request is a \emph{blueprint contract}: a grade (1--12), a CCSS ELA standard (e.g., \texttt{RL.9-10.3}) with its official description, an item type---multiple choice (MCQ), multiple select (MSQ), or fill-in-the-blank (cloze)---and a difficulty label (easy/medium/hard).
The model must emit a strict JSON object containing the question (with student-facing directions), the answer key---for cloze, an \emph{exhaustive array of all acceptable answers}---answer options where applicable, and a diagnostic explanation.

All items are scored by a commercial multi-dimensional expert evaluator (version 2.3.3; referred to throughout as the \emph{deployed evaluator}), an LLM-based judge that grades against 100+ diagnostic criteria aggregated into multi-dimensional scores (factual accuracy, clarity, difficulty alignment, standard alignment, etc.); the overall score is bottlenecked by the weakest dimension.
Following the platform's deployment criterion, an item \emph{passes} iff its overall score exceeds $0.85$.
The judge is independent of, and different from, every generator model we test; we return to judge-circularity risks in \S\ref{sec:limitations}.

\subsection{Stage 1: The Architect}
\label{sec:stage1}
The system prompt instantiates a \emph{Chief Curriculum Architect} persona and carries the invariant scaffolding: four ``golden principles'' (diagnostic validity, pedagogical distractors, cognitive alignment, construct integrity), domain constraint blocks (e.g., the separation of decoding standards from vocabulary standards; prosody rather than comprehension for fluency standards), linguistic technical rules (phonetics, letter anatomy, locale), and a 12-item quality-gate checklist.
Difficulty is operationalized by a grade-specific rubric anchored in Bloom's taxonomy---e.g., a Grade-3 \emph{hard} item must require evaluative point-of-view reasoning, while a Grade-12 \emph{hard} item must synthesize multiple interpretations---which prevents the degenerate strategy of raising difficulty via obscure vocabulary.
Generation follows a mandatory four-phase protocol executed inside the model's reasoning chain: \textsc{analyze} (restate the standard; enumerate 3--4 misconceptions), \textsc{blueprint} (map each distractor to a named misconception, or select the cloze ``golden token''), \textsc{draft}, and \textsc{validate} against the gate checklist before emitting JSON.

Two type-specific mechanisms are worth noting.
For MCQ/MSQ, each distractor must be justified by the schema \emph{``a student who chose X probably thinks \_\_ because \_\_''}, with grammatical parallelism and $\pm20\%$ option-length balance enforced.
For cloze, the protocol enforces \emph{surgical deconstruction}: the model first writes a complete, blank-free analytical passage (the ``source of truth''), then masks the single pivotal token, then performs \emph{hostile pruning}---paraphrasing away any synonym or definition of the masked token that remains visible (``mirror cues''), so the student must infer the answer from logic rather than lexical echo.

\subsection{Stage 2: The Adversary}
\label{sec:stage2}
Rather than an independent reviewer call, Stage 2 exploits \emph{conversational continuity} (the prompt is abridged in Appendix~\ref{app:attacks}): the full Stage-1 context---system prompt, user contract, draft JSON, and (where the API exposes it) the reasoning trace---is extended with a new user turn that re-roles the model as a \emph{hostile adversarial reviewer} instructed to find flaws, not to approve.
For cloze items the reviewer executes five named attacks:
(1)~\emph{answer-array precision}: substitute every key entry into the blank; delete any entry that shifts meaning, register, or part of speech;
(2)~\emph{answer-array completeness}: nominate 3--5 terms a subject-matter expert might write and add any that survive both grammatical and semantic substitution;
(3)~\emph{collocational-cue check}: read the 2--3 tokens before the blank as a phrase---if an average speaker could complete it without domain knowledge, restructure;
(4)~\emph{self-consistency}: verify the passage does not already answer the question or leak the key;
(5)~\emph{explanation audit}: recount letter counts, challenge uniqueness claims, and downgrade deductive modality (``proves'') to inductive (``suggests'') where unwarranted.
MCQ/MSQ items receive a lighter five-point gate (mirror audit, alignment, distractor logic, fact check, JSON integrity).
The reviewer outputs corrected JSON; on failure the pipeline falls back to the Stage-1 draft.
This design follows the evidence that self-correction needs external, specific criteria \citep{huang2024cannot,gou2024critic}: the attacks encode \emph{how} to break an item, not merely an exhortation to review it.

\subsection{Closing the loop: mining evaluator feedback}
\label{sec:loop}
Every evaluator verdict is recycled through two channels.

\paragraph{Bi-directional exemplars.}
Scored items are routed into an \emph{accepted} pool (23{,}530 items) and a \emph{rejected} pool (2{,}992 items), organized by grade${\times}$type${\times}$difficulty.
Each prompt samples three of each; rejected exemplars carry the evaluator's reasoning and score, and the model is instructed to name the failure pattern it must avoid.
This contrastive conditioning follows \citet{gao2024contrastive} but sources negatives from the deployed system's own failures.

\paragraph{Knowledge dictionary.}
From every scored item whose feedback contains a substantive diagnosis, we extract a \emph{knowledge rule} (Appendix~\ref{app:rule} gives one): an error category (assigned by a 10-class keyword taxonomy over the feedback text: phonetics, syllabication, dictionary skills, spelling, factual, explanation, distractor quality, grammar/morphology, etymology, difficulty mismatch), the problem description, the evaluator's ground-truth correction, and a severity score.
Rules are indexed by the exact (standard~ID, item~type) pair---2{,}263 keys, 44{,}844 rules in the snapshot used for the production run---and retrieved at generation time with \emph{diversity-first severity sampling}: one rule per category first (most severe first), then remaining slots by severity, ten rules per prompt.
This turns the evaluator's past corrections into standard-specific guardrails, in the spirit of principle learning from mistakes \citep{zhang2024leap,sun2024ricp} and deployment-time feedback memory \citep{madaan2022memprompt}, but grounded in an external expert judge rather than self-critique.
The mechanism is motivated by an early finding: in the February baseline round, virtually no rejected item failed on \emph{logic}; failures concentrated in ``capillary'' knowledge (schwa identification, syllable typing, guide-word ordering) that prompt-side knowledge injection addressed directly.
This reading was taken over the same category vocabulary whose reliability we measure in \S\ref{sec:errors}, where those particular classes turn out to over-fire substantially, so we treat it as the design rationale it was rather than as a quantitative finding.

\subsection{Iterative rule accretion}
\label{sec:rounds}
MCQ/MSQ shipped with ${\sim}12$ static format rules and never needed iteration; Appendix~\ref{app:rules} excerpts the cloze families.
Cloze rules grew over four rounds of failure-driven tuning: R0 (${\sim}15$ core rules), R1 ($+$semantic-substitution tests, format coherence, factual rigor; ${\sim}25$), R2 ($+$retrieval bans, academic-term differentiation; ${\sim}35$), R3 ($+$collocational-cue ban, self-consistency re-read, explanation audit; ${\sim}42$), and a final R4 adding word-bank integrity (the answer array must be a subset of any displayed word bank) and grade-band-specific safeguards, dispatched by grade band (K--2, 3--5, 6--8, 9--12).

\subsection{Corpus and experimental setup}
\label{sec:setup}

\paragraph{Scope.}
755 CCSS ELA standards across grades 1--12; three item types; three difficulty levels.
Over the development period (2026-02-14 to 2026-03-07) the pipeline generated 65{,}735 item instances in 534 batches; 277 early development or aborted batches (19{,}775 instances) were never submitted for evaluation. The remaining batches produced 47{,}278 valid evaluator verdicts, which resolve to 43{,}227 unique scored items plus 4{,}051 repeat verdicts on re-evaluated batches. All quality statistics use one primary verdict per item; repeats serve only the judge-reliability audit (\S\ref{sec:reliability}). Table~\ref{tab:attrition} itemizes the accounting.

\begin{table}[t]
\centering\small
\setlength{\tabcolsep}{6pt}
\begin{tabular}{lr}
\toprule
Corpus accounting & Count\\
\midrule
Generated item instances (534 batches) & 65{,}735\\
\quad in 277 early/aborted batches never evaluated & 19{,}775\\
Valid evaluator verdicts & 47{,}278\\
\quad unique items scored (primary verdicts; all quality statistics) & 43{,}227\\
\quad repeat verdicts (judge-reliability audit only, \S\ref{sec:reliability}) & 4{,}051\\
\bottomrule
\end{tabular}
\caption{Corpus accounting. A verdict is one item${\times}$evaluation pair. An item's \emph{primary} verdict comes from the evaluation of its batch carrying the most numeric scores, ties to the earliest; further verdicts on the same items are repeats, used only in \S\ref{sec:reliability}. Records whose score field reads \texttt{Error} are evaluator failures rather than quality failures (519 of 48{,}334) and are excluded throughout.}
\label{tab:attrition}
\end{table}

\paragraph{Models.}
Ten LLMs served as generators via a unified API layer: Claude Opus 4.6 and 4.5, Gemini 3.1 Pro, 3.1 Flash-Lite, and 3 Flash, GPT-5.4, GPT-OSS-120B, Kimi K2.5, GLM-5, and MiniMax M2.5; the logs also contain one small run (288 cloze items) whose generator the logs do not identify, excluded from Table~\ref{tab:models} and from every per-model claim.
All ran with extended reasoning enabled, a 20{,}480-token generation budget, and provider-default temperature.
They did \emph{not} all run the same generation protocol: the corpus records two, the two-stage protocol of \S\ref{sec:pipeline} and an earlier single-pass knowledge-grounded baseline, and three models were run exclusively under the latter (Table~\ref{tab:models}, Appendix~\ref{app:method}).
The matched comparison of Table~\ref{tab:g12} is 100\% two-stage for all five of its models; the aggregate table is not, which is one reason we treat it as observational.
The judge (evaluator version 2.3.3) and the pass criterion ($s>0.85$) were held fixed throughout.

\paragraph{Comparisons.}
Because the pipeline itself evolved during development, aggregate per-model numbers (Table~\ref{tab:models}) are observational.
For controlled claims we rely on (i)~the \emph{matched run}, in which five models generated Grade-12 cloze items under the same mature rule set, standards, and judge (Table~\ref{tab:g12}), and (ii)~within-model trajectories across rounds (Figure~\ref{fig:trajectory}).

\section{Results}
\label{sec:results}

\subsection{Full-curriculum production quality}
\label{sec:production}

\begin{table}[t]
\centering\small
\setlength{\tabcolsep}{4.2pt}
\begin{tabular}{lrrrrrr}
\toprule
& & & & \multicolumn{3}{c}{Pass rate by difficulty (\%)}\\
\cmidrule(lr){5-7}
Type & $n$ & Pass\% & Score & Easy & Med. & Hard\\
\midrule
MCQ     & 2{,}644  & \best{99.13} & 0.953 & 99.4 & 98.9 & 99.1\\
MSQ     & 2{,}652  & 98.38 & 0.952 & 98.2 & 98.6 & 98.3\\
Cloze   & 3{,}778  & 96.48 & 0.947 & 96.5 & 97.5 & 95.4\\
\midrule
All     & 9{,}074  & 97.81 & 0.950 & 97.9 & 98.2 & 97.4\\
\bottomrule
\end{tabular}
\caption{Final production run (Claude Opus 4.6): expert-evaluator pass rate (score $>0.85$) and mean score over 755 standards, grades 1--12.}
\label{tab:production}
\end{table}

\begin{figure}[t]
\centering
\includegraphics[width=0.62\textwidth]{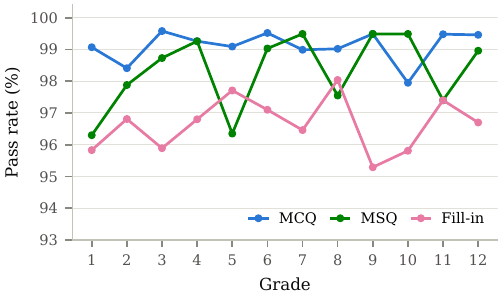}
\caption{Production pass rate by grade and item type. All 36 grade${\times}$type cells lie within a 4.3-point band (95.3--99.6\%); no grade-level degradation is observed, and hard items track easy items closely (Table~\ref{tab:production}).}
\label{fig:grade}
\end{figure}

Table~\ref{tab:production} and Figure~\ref{fig:grade} summarize the production run; Table~\ref{tab:pergrade} in Appendix~\ref{app:grades} gives the per-grade breakdown and Appendix~\ref{app:composition} the run's composition.
Three properties matter for deployment.
\emph{Level}: 97.8\% of 9{,}074 items pass an evaluator whose criteria encode professional item-writing standards.
This is a pass rate against one evaluator's criteria, not a measured defect rate: independent judges applying the same accept/reject definition agree on MCQ but reject substantially more cloze items than the deployed evaluator does (\S\ref{sec:indep}), so the level licenses a claim about throughput past this gate, not about how much human review the output still needs.
\emph{Flatness across the curriculum}: all 36 grade$\times$type cells lie within a 4.3-point band (95.3--99.6\%), and difficulty levels are indistinguishable in evaluator-judged terms (easy 97.9\%, hard 97.4\%; two-proportion $p{=}0.20$)---evidence that the grade-banded rubrics and rule dispatch neutralized evaluator-judged difficulty miscalibration at the extremes of the grade range; whether \emph{empirical} difficulty matches the labels is a psychometric question that requires field testing with student responses (\S\ref{sec:limitations}).
\emph{Residual asymmetry}: even at the top of the capability range and after all tuning, cloze remains 2.6 points below MCQ (96.5\% vs.\ 99.1\%)---the residue of the structural gap analyzed below.

\subsection{The closed-set/open-set asymmetry}
\label{sec:asymmetry}

\begin{table}[t]
\centering\small
\setlength{\tabcolsep}{3.4pt}
\begin{tabular}{lcrrr}
\toprule
Model & 2-pass & MCQ & MSQ & Cloze\\
\midrule
Claude Opus 4.6      & 100\% & \best{99.1} {\scriptsize(2{,}644)} & \best{98.4} {\scriptsize(2{,}652)} & \best{96.3} {\scriptsize(3{,}969)}\\
Claude Opus 4.5      & 100\% & --- & --- & 93.2 {\scriptsize(191)}\\
Gemini 3.1 Pro       & 100\% & 98.7 {\scriptsize(3{,}373)} & 98.3 {\scriptsize(3{,}641)} & 87.7 {\scriptsize(6{,}855)}\\
GLM-5                & 0\%   & 95.3 {\scriptsize(278)} & 97.2 {\scriptsize(36)} & 86.5 {\scriptsize(408)}\\
Gemini 3 Flash       & 100\% & --- & --- & 86.8 {\scriptsize(83)}\\
GPT-5.4              & 100\% & --- & --- & 82.8 {\scriptsize(192)}\\
Kimi K2.5            & 27\%  & 95.2 {\scriptsize(8{,}033)} & 91.7 {\scriptsize(4{,}224)} & 78.4 {\scriptsize(5{,}093)}\\
GPT-OSS-120B         & 67\%  & --- & --- & 74.0 {\scriptsize(254)}\\
Gemini 3.1 Fl.-Lite  & 100\% & --- & 84.6 {\scriptsize(286)} & 65.0 {\scriptsize(623)}\\
MiniMax M2.5         & 0\%   & --- & --- & 59.6 {\scriptsize(104)}\\
\bottomrule
\end{tabular}
\caption{Aggregate pass rate (\%) by model and item type over all evaluated runs ($n$ in parentheses). \textbf{Observational, not a controlled comparison}: rows pool pipeline maturity stages and two generation methods. ``2-pass'' is the share of each model's items produced by the two-stage protocol of \S\ref{sec:pipeline}, the remainder by a single-pass baseline; method and model are partly collinear, so the spread is not attributable to capability alone (Appendix~\ref{app:method}). The controlled comparison is Table~\ref{tab:g12}. The MCQ--cloze gap appears for every model measured on both.}
\label{tab:models}
\end{table}

\begin{figure}[t]
\centering
\includegraphics[width=0.62\textwidth]{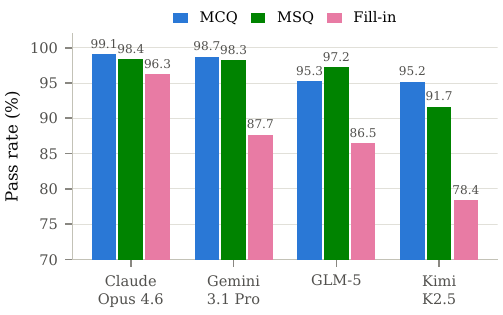}
\caption{Pass rate by item type for the four models with all three types measured. MCQ/MSQ cluster at 92--99\% while cloze spans 78--96\%. Every model loses ground on cloze---Claude Opus 4.6 $-2.8$, GLM-5 $-8.8$, Gemini 3.1 Pro $-11.0$, Kimi K2.5 $-16.8$ points against its own MCQ rate---largest for the weakest model and smallest for the strongest, though not strictly monotone: GLM-5 is below Gemini 3.1 Pro on both formats yet has the smaller gap, on the smallest sample here and entirely single-pass (Table~\ref{tab:models}).}
\label{fig:gap}
\end{figure}

\begin{table}[t]
\centering\small
\setlength{\tabcolsep}{4.5pt}
\begin{tabular}{lrrr}
\toprule
Model & $n$ & Pass\% & 95\% CI\\
\midrule
Claude Opus 4.6 & 576   & \best{96.7} & [94.9, 97.9]\\
Claude Opus 4.5 & 191   & 93.2 & [88.7, 96.0]\\
Gemini 3.1 Pro  & 1{,}055 & 90.2 & [88.3, 91.9]\\
Kimi K2.5       & 192   & 88.5 & [83.3, 92.3]\\
GPT-5.4         & 192   & 82.8 & [76.8, 87.5]\\
\bottomrule
\end{tabular}
\caption{Matched run: Grade-12 cloze generation under the identical mature rule set, standards, and judge. Wilson 95\% intervals. Claude Opus 4.6 is separated from all non-Claude models: 13.9 points span the five models here, against a 3.9-point span (95.2--99.1\%) across the four models with MCQ measurements (Table~\ref{tab:models}). Two of these five models, Claude Opus 4.5 and GPT-5.4, generated no MCQ items at all, so the MCQ comparison is over an overlapping but not identical model set.}
\label{tab:g12}
\end{table}

Table~\ref{tab:models} and Figure~\ref{fig:gap} show the asymmetry that motivates our analysis.
Every model that generated both formats passes MCQ at 95--99\% but drops 2.8--16.8 points on cloze; the drop is ordered by model capability.
The matched Grade-12 run (Table~\ref{tab:g12}) makes the comparison controlled: under the identical prompt, standards, and judge, cloze pass rates tier the five models across a 13.9-point range, with non-overlapping confidence intervals between the top model and all non-Claude systems.
The contrast with MCQ is one of \emph{spread}, not of statistical indistinguishability: the four models with MCQ measurements occupy a 3.9-point band (95.2--99.1\%) against 13.9 points on matched cloze, a spread 3.6 times wider.
Within the MCQ band the models are in fact separable given the sample sizes involved---Kimi K2.5 at 95.19\% ($n{=}8{,}033$) sits well below Claude Opus 4.6 at 99.13\% ($n{=}2{,}644$), $p<10^{-15}$---so the claim is that closed-set generation compresses the capability ordering into a few points near ceiling, not that it erases it.
Two of the five matched models, Claude Opus 4.5 and GPT-5.4, have no MCQ items in the corpus, so the two bands are measured over overlapping rather than identical model sets.
Cloze generation, in other words, functions as a discriminative stress test: it exposes a capability ordering that saturated closed-set formats conceal, consistent with reports that closed-form benchmarks overestimate model competence \citep{balepur2024artifacts,chandak2025answermatching}.

\paragraph{Flatness across grades.}

The production run holds every grade${\times}$type cell inside a 4.3-point band (Figure~\ref{fig:grade}), which invites reading curriculum flatness as a property of the \emph{pipeline}. It is not automatic. Table~\ref{tab:gradient} reports a single earlier run of Gemini 3.1 Pro, where one generator, one rule set, one set of standards and one judge are held fixed and only format and grade vary.

Across grades the open-set format moves 25.1 points while the closed-set formats move 2.0 and 2.6. The sharpest cell is Grade~11, where all three formats were generated in equal number ($n{=}192$ each): 99.5\% on MCQ and 97.9\% on MSQ against 69.8\% on cloze. Nothing distinguishes those three measurements except the format, so a grade gradient of this magnitude is a property of open-set key construction rather than of grade difficulty in general.

This does not attribute the gradient to capability. That run used an earlier rule generation than the production run, and rule maturity moves this generator materially---under the same model, Grade-12 cloze rises from 82.9\% to 90.2\% as rules accrete---so its distance from the production run mixes capability with scaffolding, and the corpus holds no mature-rule multi-grade cloze run for a second generator to separate them. The model effect is isolated instead by the matched run of Table~\ref{tab:g12}: under one rule set, 96.7\% against 90.2\%, a 6.5-point gap, real and far smaller than the 25 points the open-set gradient spans.

\begin{table}[t]
\centering\small
\setlength{\tabcolsep}{5pt}
\begin{tabular}{llrrr}
\toprule
Format & Grades measured & $n$ & Pass rate (\%) & Span\\
\midrule
MCQ (closed-set)     & 9--12          & 771     & 97.4--99.5 & 2.0\\
MSQ (closed-set)     & 1, 10--12      & 792     & 95.3--97.9 & 2.6\\
Cloze (open-set)     & 1--2, 4, 7--11 & 2{,}570 & 69.8--94.8 & \best{25.1}\\
\midrule
\multicolumn{5}{l}{Grade 11, all three formats, $n{=}192$ each: MCQ 99.5, MSQ 97.9, cloze 69.8}\\
\bottomrule
\end{tabular}
\caption{One run of Gemini 3.1 Pro under a fixed rule generation, by format. Generator, rule set, standards and judge are constant; only format and grade vary. Coverage differs by format because the run was not exhaustive, so grades measured are listed rather than interpolated. The open-set format spans an order of magnitude more than either closed-set format across grades, and at the one grade where all three were generated in equal number the MCQ--cloze gap is 29.7 points. This run predates the production run's mature rule set, so it is not comparable to Table~\ref{tab:production} in level.}
\label{tab:gradient}
\end{table}

The difficulty label behaves the same way (Table~\ref{tab:difficulty}). Gemini's cloze pass rate falls 10.8 points from easy to hard items, while its MCQ falls 1.9 and its MSQ is flat---the label ``hard'' costs accuracy only where the answer boundary is open-set. Both observations sharpen the asymmetry along the axes we can vary cleanly: within a fixed run the closed-set formats are flat across grade and across the difficulty label, and the open-set format is not. Whether open-set flatness is reached by capability or by scaffolding is not separable in this corpus, as noted above; what Table~\ref{tab:g12} does show is that a 6.5-point model gap survives when the rule set is held fixed.

\begin{table}[t]
\centering\small
\setlength{\tabcolsep}{6pt}
\begin{tabular}{lrrrr}
\toprule
Type & Easy & Medium & Hard & Hard$-$Easy\\
\midrule
Cloze & 91.6 & 90.8 & \best{80.8} & $-10.8$\\
MCQ   & 99.5 & 99.1 & 97.6 & $-1.9$\\
MSQ   & 98.4 & 98.7 & 97.9 & $-0.5$\\
\bottomrule
\end{tabular}
\caption{Gemini 3.1 Pro pass rate (\%) by difficulty and type, aggregated over all grades ($n{=}2.3$K per cloze cell). One model, one table: only the open-set format degrades with the difficulty label, while the closed-set formats are difficulty-robust. This is a within-model comparison across formats, so it does not speak to how difficulty-robustness varies between models.}
\label{tab:difficulty}
\end{table}

\subsection{How much should we trust the judge?}
\label{sec:reliability}
Because every number above is a verdict from a single LLM-based judge, we audit its stability using the 4{,}051 repeat verdicts present in the corpus: 2{,}778 items (998 MCQ, 1{,}372 MSQ, 408 cloze) were independently re-evaluated two to three times during routine re-runs, all under the same judge version.
Within-item score variation is small (mean per-item score SD $\leq 0.009$ on the $[0,1]$ scale), the binary pass/fail verdict flips between repeats for 6.2\% of MCQ, 2.6\% of MSQ, and 6.6\% of cloze items, and batch-level pass rates move by 1.8 points on average between repeat evaluations of the same batch (max 8.0 points, on one 50-item batch).
Two consequences follow.
First, single-verdict pass rates carry roughly $\pm$1--2 points of judge noise, so small between-condition differences should not be over-read.
Second---and central to our argument---flip rates for cloze (6.6\%) and MCQ (6.2\%) are nearly identical, so the 2.8--16.8-point format gap of \S\ref{sec:asymmetry} cannot be an artifact of noisier cloze judging; the most stable format under repeats (MSQ, 2.6\%) sits \emph{between} the two in pass rate.

Two properties of the repeat set qualify this audit.
It is not a random subsample: re-runs were triggered operationally, and 10 of the 19 re-evaluated batches were re-run after their first evaluation returned errors on some items.
An item enters the audit only if it was scored in every evaluation of its batch, which for a partly-failed first evaluation means only the subset that succeeded there.
Relaxing that to ``scored in at least two evaluations'' adds 280 items and moves the cloze flip rate to 8.2\% while MCQ stays at 6.2\%, so the cloze-versus-MCQ flip comparison is worth 0.4 or 1.9 points depending on the pairing rule---either way an order of magnitude below the format gap it is used to rule out.
This audit bounds the judge's random error; systematic bias shared across repeats (e.g., a stable preference for particular phrasings) is invisible to it, which is what the independent-judge check below addresses directly.

\paragraph{Independent judges.}
\label{sec:indep}
The repeat audit bounds random error but not bias the judge holds stably, so we re-scored a sample with judges sharing no vendor with the generator or with each other.
We drew 390 items: 100 per item type from the matched production run (the \emph{asymmetry} stratum), plus 30 deployed failures per type (the \emph{agreement} stratum, oversampled because the production failure rate is only 2--4\% and $\kappa$ is otherwise inestimable).
Three judges---Gemini 3.1 Pro, GPT-5.1, and open-weights Qwen3-235B---scored every item blind to the deployed verdict, the deployed score, and the stratum (Appendix~\ref{app:indep} details the sample and rubric), under our own accept/reject definition (key wrong, key incomplete, ambiguous stem, off-standard, off-grade) rather than the deployed evaluator's criteria, so agreement is not manufactured by handing a judge the same rubric.

\begin{table}[tb]
\centering\small
\setlength{\tabcolsep}{4pt}
\begin{tabular}{lrrrr}
\toprule
Judge & MCQ & MSQ & Cloze & Cloze$-$MCQ\\
\midrule
Gemini 3.1 Pro   & 96.0 & 96.0 & 67.3 & $-28.7$\\
GPT-5.1          & 97.0 & 95.0 & 87.0 & $-10.0$\\
Qwen3-235B       & 100.0 & 100.0 & 99.0 & $-1.0$\\
\midrule
Majority of the three & 99.0 & 100.0 & 91.0 & $-8.0$\\
Deployed evaluator & 98.0 & 100.0 & 96.0 & $-2.0$\\
\bottomrule
\end{tabular}
\caption{Accept rate (\%) on the asymmetry stratum, 100 items per type, judges blind to the deployed verdict. Denominators are 100 except where a judge returned an unusable reply: Gemini 3.1 Pro scored 99 MSQ and 98 cloze items. The format ordering reproduces under every judge, and the two frontier judges recover a \emph{larger} cloze deficit than the deployed evaluator does. Two-proportion tests on Cloze$-$MCQ: $p{=}1.7{\times}10^{-7}$ (Gemini), $0.009$ (GPT-5.1), $0.32$ (Qwen3), $0.009$ (majority), $0.41$ (deployed).}
\label{tab:indep}
\end{table}

Three things follow (Table~\ref{tab:indep}).
\emph{The asymmetry is not the deployed judge's artifact.} Cloze is the lowest-accepted format for all three judges, significantly so for both frontier judges and for their majority vote, and the effect is \emph{larger} outside the deployed evaluator than inside it ($-28.7$ and $-10.0$ points vs.\ $-2.0$). Restricting to items the deployed evaluator \emph{passed}, Gemini 3.1 Pro and GPT-5.1 concur on rejecting 9.9\% of cloze against 1.0\% of MCQ and 0.0\% of MSQ---a tenfold format difference in concordant residual defects.
\emph{The residual defect is the one the mechanism predicts.} Of the cloze items they reject that the deployed evaluator passed, answer-key incompleteness is the dominant charge (GPT-5.1 12 of 13; Gemini 18 of 31), while on MCQ it is rare. Judges given no knowledge of our thesis converge on open-set key omission (\S\ref{sec:analysis}) as the failure that survives the pipeline.
\emph{The absolute level, however, is judge-relative.} Agreement with the deployed accept/reject binary is 77--79\%, and Cohen's $\kappa$ falls from 0.33--0.43 on the fail-oversampled sample to 0.13 once reweighted to production prevalence. This is weak agreement, and it is the honest bound on what ``97.8\% pass'' certifies: throughput past one gate, not a measured defect rate.
The bound is not specific to the deployed evaluator. The judges agree with \emph{each other} no better ($\kappa{=}0.40$ Gemini--GPT-5.1, $0.10$ and $0.12$ for the pairs involving Qwen3), so item acceptability is a low-agreement judgment among competent judges rather than a fact the deployed evaluator alone gets wrong.
Qwen3-235B accepts 99.7\% of the asymmetry stratum and recovers 8.3\% of deployed failures: as a judge it reproduces the verdict form without the accept/reject boundary, the same dissociation the distillation study of \S\ref{sec:distill} induces deliberately (\S\ref{sec:distill}).

These are model judges, not human experts, and they do not establish ground truth; the human validation and field testing of \S\ref{sec:limitations} remain the necessary next step.
What they do establish is that the format asymmetry survives a change of judge, vendor, and rubric, and that the deployed evaluator is the most lenient of the four on exactly the format our analysis predicts is hardest.

\subsection{Prompt optimization plateaus, and the error mass trades direction}
\label{sec:pareto-results}

\begin{figure}[t]
\centering
\includegraphics[width=0.62\textwidth]{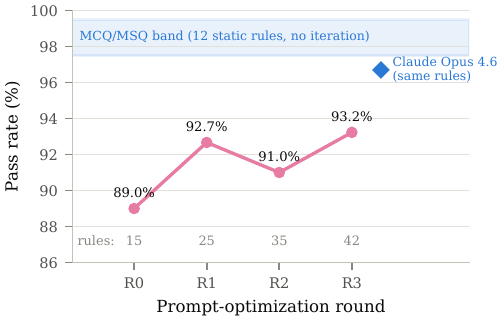}
\caption{Grade-12 cloze pass rate across optimization rounds (Gemini 3.1 Pro) as rules accrete from 15 to 42. Four rounds buy $+4.2$ points and plateau ${\sim}5$ points below the MCQ/MSQ band, which required no iteration; swapping the generator to Claude Opus 4.6 under the same rules recovers most of the residual gap.}
\label{fig:trajectory}
\end{figure}

\begin{figure}[t]
\centering
\includegraphics[width=0.45\textwidth]{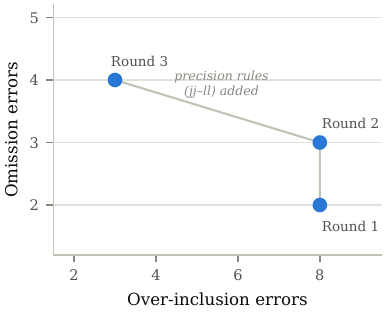}
\caption{Answer-key error counts across rounds (Grade-12 audit set). Precision-oriented rules added after R2 cut over-inclusion from 8 to 3 cases but raised omissions from 2--3 to 4, leaving total boundary-error mass roughly constant. Counts are small---10--12 failing items per round, one grade, one generator---so this is the shape of a precision--recall trade rather than a measured frontier.}
\label{fig:pareto}
\end{figure}

Figure~\ref{fig:trajectory} traces the controlled Grade-12 trajectory for Gemini 3.1 Pro: 89\% at R0, 92.7\% at R1, a \emph{regression} to 91\% at R2 despite ten additional rules, 93.2\% at R3---then a plateau, five points short of the MCQ band that the same model reaches with a twelfth of the rule mass and zero iteration.
The R2 dip coincides with the largest single-round rule addition, consistent with documented constraint-interference effects \citep{jiang2024followbench,wen2024complexbench}.
The full-scale (all-grades) trajectory is consistent: 86.2${\to}$88.0${\to}$88.4${\to}$90.2\% across four successive pipeline days.
Rule scaffolding does transfer across models---Kimi K2.5's Grade-12 cloze rate climbed from 68.6\% (Feb.\ 27) to 88.5\% (Mar.\ 6) under the accreting rule set, a $+20$-point gain---but no amount of scaffolding closed the gap for any model below the frontier tier.

Figure~\ref{fig:pareto} shows \emph{why} iteration stalls.
Manual audit of every failing Grade-12 cloze item per round found the two answer-key error directions trading places: rounds that added precision rules (``accept only the standard academic label'') cut over-inclusion (8${\to}$3 cases) while omissions rose (2${\to}$4); the over-inclusion/omission ratio inverted from 4.0 to 0.75 while the \emph{total} boundary error mass stayed roughly constant.
This audit is small---10--12 failing items per round, one grade, one generator---so we read it as suggestive evidence of a Pareto frontier rather than proof; it is, however, consistent with the plateau in Figure~\ref{fig:trajectory} and with the constraint-saturation literature \citep{jiang2024followbench,jaroslawicz2025ifscale}.
On this reading prompt rules act as regularizers that slide the operating point along a precision--recall trade-off for the answer-set boundary rather than supplying the missing discriminative capability; the audit is too small to establish that frontier, and we offer it as the interpretation most consistent with the plateau.
Notably, the highest-capability generator moved the frontier itself (Claude Opus 4.6: 96.7\% under the same rules), indicating the ceiling is a joint property of task structure and model capability, not of the rule set.

\subsection{What still fails: error taxonomy}
\label{sec:errors}

\begin{table}[t]
\centering\small
\setlength{\tabcolsep}{4.5pt}
\begin{tabular}{lrrr}
\toprule
Error category & Cloze & MCQ & MSQ\\
\midrule
Difficulty mismatch      & \best{22.7} & 7.6  & 8.2\\
Explanation defects      & 20.9 & 19.4 & 20.1\\
Factual errors           & 13.8 & 16.7 & 17.3\\
Phonetics claims         & 10.1 & 8.5  & 8.7\\
Grammar/morphology       & 9.0  & 4.9  & 3.0\\
Distractor quality       & 7.6  & \best{32.3} & \best{33.3}\\
Other (incl.\ key bounds) & 15.9 & 10.6 & 9.4\\
\bottomrule
\end{tabular}
\caption{Distribution of error-category labels (\%) over all 3{,}556 failing items, via a 10-class keyword taxonomy applied to judge feedback (classes under 3\% merged into ``Other''). \textbf{These shares are instrument-limited}: validated against independent re-labelling of 450 failing items, the keyword rule agrees only weakly ($\kappa{=}0.27$) and over-fires wherever the feedback merely mentions a category's vocabulary---per-class precision runs 0.03--0.06 for \textsc{spelling} and \textsc{phonetics} and 0.23 for \textsc{distractor\_quality}, against 0.76 for \textsc{factual\_error} and 0.69 for \textsc{explanation\_error}. Only the cloze difficulty-mismatch concentration survives the check; see the text.}
\label{tab:errors}
\end{table}

Table~\ref{tab:errors} profiles the 3{,}556 items that failed across all runs, and we validated the labelling instrument before reading it: an independent model re-labelled 450 failing items (150 per format) from the same feedback text under the same label definitions (Appendix~\ref{app:taxonomy}).
Agreement with the keyword rule is poor ($\kappa{=}0.27$ on the primary label; mean Jaccard 0.38), and the disagreement is systematic rather than random---substring rules fire on a category whenever the feedback \emph{mentions} its vocabulary, which in ELA feedback it constantly does.
Of the three format patterns the profile appears to show, one survives the check.
\emph{Cloze failures concentrate in difficulty mismatch}: 36.0\% of failing cloze items by keyword rule and 36.7\% under independent labelling, against 17.3\% and 12.7\% for MCQ---the format contrast holds and widens under the stricter labeller. These are overwhelmingly items judged easier than labeled because the blank was solvable by retrieval or collocation (\S\ref{sec:collocation}).
The apparent concentration of MCQ/MSQ failures in \emph{distractor quality} does not survive: the keyword rule assigns that label to 90.7\% of failing MCQ items but the independent labeller to 22.0\%, and the format gap reverses in sign ($+80.7$ points keyword vs.\ $-7.3$ points independent), because evaluator feedback on an MCQ nearly always discusses the distractors whatever the actual defect.
Nor does the apparent long tail of phonetics and reference-skill failures: 17--22\% by keyword rule across formats versus 0.7--2\% independently, so we withdraw the inference that these domains specifically warrant retrieval or tool grounding.
The independent labeller also assigns no fitting class to 21--43\% of failures (keyword rule: 0.7--21\%), indicating the ten classes under-cover the failure space.

\subsection{The asymmetry recurs when distilling the judge}
\label{sec:distill}

If open-set discrimination is the binding constraint, it should resist not only \emph{generating} boundary-correct items but also \emph{learning to judge} them.
We test this directly by distilling the expert evaluator itself.
Each scored item is paired with its deployed-evaluator verdict (nine dimension scores, a pass/fail decision at $0.85$, and a free-text diagnosis); we split at the batch level---holding out 8\% of batches (3{,}593 verdicts) so no evaluation item shares a batch with training---and LoRA-fine-tune a Qwen2.5-14B-Instruct student (rank 32) to reproduce the verdict from the item alone.
Judging item quality is itself the open-set task of \S\ref{sec:formal}: deciding whether an answer key is complete and exclusive, and whether difficulty is calibrated, is a discrimination call, not a construction call.
The teacher supplies ${\sim}40$K real discrimination labels---if discrimination transferred by imitation, this should suffice.

\begin{figure}[t]
\centering
\includegraphics[width=0.66\linewidth]{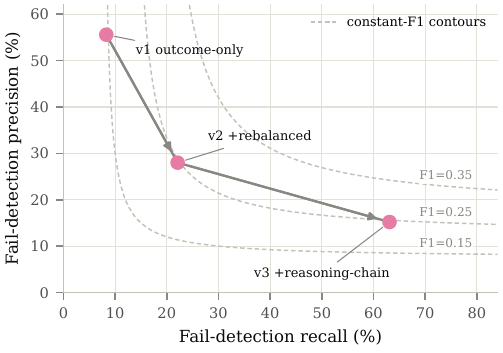}
\caption{Fail-detection precision vs.\ recall for the three distillation recipes; dashed curves are loci of constant F1. Fail-oversampling does buy discrimination once (v1${\to}$v2 lifts F1 from 0.14 to 0.25, crossing contours), but the richer reasoning-chain supervision (v2${\to}$v3) then moves the student \emph{along} the F1${\approx}0.25$ contour rather than past it, trading precision for recall at constant F1. Arrows give the direction of the recipe sequence.}
\label{fig:frontier}
\end{figure}

It does not.
Table~\ref{tab:distill} traces three recipes.
Outcome-only distillation (v1) reproduces the \emph{surface} of the judge perfectly---every output is a well-formed multi-dimensional verdict---but collapses to the majority class: it predicts ``pass'' almost always, matching the 93.3\% base rate at 8.3\% fail-recall.
Oversampling failing items $4\times$ (v2) and distilling the evaluator's full reasoning trace rather than only its verdict (v3) raise fail-recall to 22.1\% and 63.1\%, but F1 is flat past v2 ($.25{\to}.25$) and pass/fail agreement \emph{falls below} the trivial always-pass baseline (93.3\%) as recall climbs, because the student now over-flags: precision drops from 55.6\% to 15.2\%.
Figure~\ref{fig:frontier} plots the three recipes in recall--precision space. The first step does buy discrimination---v1${\to}$v2 lifts F1 from 0.14 to 0.25---but the second does not: v2 and v3 sit on the same F1${\approx}0.25$ contour, sliding from over-\emph{omission} of ``fail'' to over-\emph{inclusion} of it, the same precision--recall trade of Figure~\ref{fig:pareto} now surfacing in the evaluation task rather than the generation task.
No recipe reaches the teacher's own repeat-verdict consistency band (93.4--97.4\%, \S\ref{sec:reliability}) while actually discriminating.

\begin{table}[t]
\centering\small
\setlength{\tabcolsep}{4.5pt}
\begin{tabular}{lrrrrrrrr}
\toprule
 & \multicolumn{5}{c}{Overall} & \multicolumn{3}{c}{Agreement by type}\\
\cmidrule(lr){2-6}\cmidrule(lr){7-9}
Distillation recipe & Agree.\ & Rec.\ & Prec.\ & F1 & $r$ & Cloze & MCQ & MSQ\\
\midrule
Trivial ``always pass''      & 93.3 & 0.0  & ---  & --- & ---  & ---  & ---  & ---\\
v1 \; outcome-only           & 93.4 & 8.3  & \best{55.6} & .14 & .19 & 89.6 & \best{96.8} & 93.2\\
v2 \; {+}\,fail-rebalanced   & 91.0 & 22.1 & 28.0 & .25 & .17 & 86.2 & 95.8 & 90.0\\
v3 \; {+}\,reasoning-chain   & 73.9 & \best{63.1} & 15.2 & .25 & .24 & \best{58.6} & 89.0 & 70.8\\
\midrule
Teacher self-consistency     & 93.4--97.4 & --- & --- & --- & --- & --- & --- & ---\\
\bottomrule
\end{tabular}
\caption{Distilling the deployed evaluator into a Qwen2.5-14B student, on 3{,}593 held-out verdicts (\%; fail-detection treats the 6.7\% of items the teacher fails as the positive class; $r$ is Pearson correlation with the teacher's continuous score). Raising fail-recall across recipes (8.3${\to}$63.1) leaves F1 flat and drives agreement \emph{below} the always-pass baseline. The per-type columns show the asymmetry recurring at the meta-level: the distilled judge agrees least, and degrades most (89.6${\to}$58.6), on the open-set cloze format---judging cloze is harder than judging MCQ, just as generating it is.}
\label{tab:distill}
\end{table}

This is a second, independent instance of the paper's thesis, and it isolates \emph{mechanism~(ii)} of \S\ref{sec:mechanisms}: the generative surface of judging transfers readily; the discriminative boundary does not, even from ${\sim}40$K labeled verdicts of a strong teacher.
The per-type breakdown (Table~\ref{tab:distill}, right) is itself the asymmetry at one remove: agreement is lowest on cloze for every recipe and collapses fastest there (89.6${\to}$58.6 vs.\ 96.8${\to}$89.0 for MCQ), so the open-set format is the hardest to \emph{judge}, not only to generate.
It also bears on judge circularity (\S\ref{sec:limitations}): the difficulty is a property of the discrimination \emph{task}---surfacing when we try to replicate the judge---not merely of the particular judge we deploy.

\emph{Caveats.}
The student is a single 14B model; the capability tiering of Table~\ref{tab:g12} predicts a larger student may move the frontier outward, so this establishes the asymmetry, not its scale-invariance.
Whether the discrimination gain is scale-emergent or frontier-bound would take distillation into a substantially larger student, which we did not run.
v3's over-flagging is partly induced by the reasoning recipe's explicit flaw-seeking instruction, so the frontier \emph{position} is recipe-dependent even if the frontier itself is not; the continuous-score correlation stays low throughout (Pearson $r{=}0.17$--$0.24$).
This is a cross-task corroboration of---not a substitute for---the generation-side mechanisms of \S\ref{sec:analysis}.

\section{Analysis and Discussion}
\label{sec:analysis}

\subsection{Closed-set vs.\ open-set generation}
\label{sec:formal}
An MCQ generator emits $(q, O, k)$: stem, option set, key.
The judge's core check is \emph{internal consistency}---exactly one option correct, distractors defensibly wrong:
\begin{equation*}
P(\text{pass}) \!=\! P\big(k \text{ correct} \wedge \forall_{i\neq k}\, o_i \text{ wrong} \,\big|\, q, O\big),
\end{equation*}
where every argument is the model's own construction.
The answer space is \emph{closed and self-referential}: the model plays both architect and inspector of a 2-bit decision space ($\log_2 4$).

A cloze generator must emit $(q, S^{*})$ where $S^{*} \subseteq V$ is the set of \emph{all} acceptable answers.
The target boundary partitions the vocabulary, $V = S_{\text{accept}} \cup S_{\text{reject}}$, and is fixed by the language and the curriculum---not by the model.
Passing requires jointly avoiding both boundary error directions:
\begin{equation*}
P(\text{pass}) = P(\text{no over-inclusion}) \times P(\text{no omission}),
\end{equation*}
a conjunction of a precision event and a recall event over a ${\sim}2^{15}$-candidate space.
The judge checks \emph{external correctness}: whether the model recovered a pre-existing semantic boundary.
This is a discrimination task embedded in a generation task.
The framing makes a prediction that can be tested away from curricula, and it holds: on constructions where $S^{*}$ is finite and mechanically decidable, the same models recover the boundary when asked to \emph{check} membership but not when asked to \emph{enumerate} it, and they recover it almost perfectly when allowed to emit the defining predicate instead of its extension \citep{chen2026judging}.

\subsection{Three mechanisms}
\label{sec:mechanisms}
The observed error modes map onto three structural properties of autoregressive decoders.

\paragraph{(i) Sequential enumeration without lookahead.}
The answer array is generated element-by-element under a causal mask: the decision to emit or withhold candidate $w_n$ conditions on $w_{<n}$ but not on candidates never sampled, and the decision to \emph{stop} enumerating is itself a local next-token prediction rather than a completeness judgment \citep{vinyals2016order,welleck2020consistency,hou2025enumeration}.
This predicts order effects and omission of valid rare synonyms---the error class that \emph{grew} as precision rules suppressed its dual (Figure~\ref{fig:pareto}).

\paragraph{(ii) Predictive, not discriminative, representations.}
Next-token training makes representations track co-occurrence: \emph{allegory} and \emph{parable} are near neighbors because they share contexts.
The key-boundary task requires the opposite judgment---whether two near neighbors are \emph{interchangeable in this context}---a discriminative distinction the training objective never directly optimizes \citep{devlin2019bert,west2024paradox}.
Empirically, LM token distributions misestimate human cloze acceptability \citep{jacobs2024cloze} and fine-grained collocate admissibility \citep{espinosa2021collocations}; our over-inclusion errors (accepting \emph{skeptical} where the text demonstrates \emph{irony}) are precisely this signature.
Distilling the evaluator itself exhibits the same mechanism from the opposite side (\S\ref{sec:distill}): ${\sim}40$K teacher verdicts transfer the verdict form but not the accept/reject boundary, so the student cannot be pushed off the precision--recall frontier by more supervision alone.

\paragraph{(iii) Uncalibrated membership thresholds.}
Emitting a key array implicitly thresholds a continuous distribution into a binary membership call per candidate.
Models are reasonably calibrated on closed formats \citep{kadavath2022know} but overconfident and unstable exactly at ambiguity boundaries \citep{xiong2024confidence,liu2023ambiguity}; the item-dependent threshold (``exact word'' vs.\ ``any valid synonym'') can only be communicated through instructions, and instruction-following itself degrades under the 40-rule loads these items carry \citep{jiang2024followbench,jaroslawicz2025ifscale}.
The observed capability tiering (Table~\ref{tab:g12}) is then expected: models differ far more in discriminative calibration under constraint load than in fluent construction, which is all MCQ demands.

\subsection{Collocational cue leakage}
\label{sec:collocation}
One error class is not merely unsolved by autoregressive generation but \emph{caused} by it.
When drafting the sentence frame, the decoder follows its probability gradient, preferentially producing high-frequency n-gram prefixes; if the blank is then placed after such a prefix (``dramatic \rule{0.8cm}{0.4pt}'', ``badge of \rule{0.8cm}{0.4pt}''), the intended answer is recoverable from collocation statistics alone, and the judge correctly downgrades the item's difficulty.
Formally, the generator maximizes $P(c)$ over frames $c$, which co-maximizes $P_{\mathrm{LM}}(w_b{=}\text{answer} \mid c)$---a systematic bias toward self-defeating blanks.
Human item writers deliberately break collocations; MCQs are immune because explicit options interrupt pattern completion (though they leak other cues; \citealp{balepur2024artifacts}).
This explains why \emph{difficulty mismatch} dominates the cloze error profile (Table~\ref{tab:errors}) and why the R3 collocational-cue ban produced the single largest error-class reduction of any rule addition.

\subsection{Design implications}
\label{sec:discussion}

\paragraph{Evaluate keys by rubric, not by enumeration.}
The binding failure mode is exhaustive key enumeration---an open-set discrimination task graded with closed-set strictness.
Ecosystems can remove the structural mismatch by scoring student responses with semantic matching against a \emph{rubric} (as argued for model evaluation by \citealp{chandak2025answermatching}) instead of demanding a verbatim-complete key array at authoring time.
Where exact keys are required, boundary determination should be externalized: a bidirectional-encoder verifier or dictionary/thesaurus tool call adjudicating each candidate's membership \citep{behnamghader2024llm2vec,gou2024critic}, rather than more prompt rules.

\paragraph{Spend prompt mass where the task is open-set.}
A dozen static rules saturate MCQ/MSQ across every model tested; the marginal return of the next thirty rules is concentrated entirely in cloze---and even there it buys a $+4.2$-point plateau, with the audited error mass trading direction rather than shrinking.
Teams should treat rule accretion as a diagnostic: when new rules start trading one error direction for its dual, the remaining gap is capability- or architecture-bound, and the lever is model choice or system redesign, not prompt engineering.

\paragraph{Cloze as a capability separator.}
MCQ generation compresses the models into a 3.9-point band near ceiling (95.2--99.1\%), while open-set key construction spreads the same tier over 13.9 points (Table~\ref{tab:g12}).
Constraint-dense cloze generation is cheap to run against any judge and, unlike saturated closed-set benchmarks, yields a graded, non-gameable ordering---we propose it as a standing probe for discriminative-under-constraint capability.

\paragraph{Feedback mining beats feedback discarding.}
The knowledge dictionary and negative exemplars convert a paid evaluation stream into standard-indexed guardrails at zero marginal labeling cost.
The mechanism is general: any generation system with a structured critic can persist (category, problem, correction, severity) tuples and retrieve them by task key \citep{zhang2024leap,madaan2022memprompt}.

\subsection{Limitations}
\label{sec:limitations}
\textbf{Judge circularity.}
Quality is defined by a single LLM-based evaluator (the deployed evaluator, version 2.3.3), and the loop optimizes against it; some gains may exploit judge idiosyncrasies rather than true quality, a known risk of judge-guided optimization \citep{zheng2023judging,panickssery2024selfrecognition}.
The judge is a fixed external commercial system distinct from all generators, its criteria encode published item-writing standards \citep{haladyna2002review}, and the asymmetry findings are consistent across ten independently trained generators.
The independent-judge cross-validation of \S\ref{sec:indep} addresses this directly and splits the question: the format asymmetry survives a change of judge, vendor, and rubric, and is larger outside the deployed evaluator than inside it, so that finding is not an artifact of this evaluator; but agreement on the accept/reject binary is weak at production prevalence ($\kappa{\approx}0.13$), so the headline \emph{level} remains judge-relative and should not be read as a measured defect rate.
That weak agreement is not evidence against the deployed evaluator specifically---the independent judges agree with each other no better---but it does mean the pipeline is optimized against a target that competent judges do not reproduce, and some gains may exploit judge idiosyncrasies rather than true quality.
Human expert validation of a stratified sample---and ultimately field testing with student response data (IRT calibration)---remains the necessary next step, and neither is reported here. We have built a pre-registered human-validation instrument for this purpose (blinded, fail-oversampled $n{=}360$ sample; two-rater workbook; Cohen's-$\kappa$ and human-adjusted-pass-rate analysis, self-tested on synthetic fixtures), so the validation can be executed and reported directly.
\textbf{Observational aggregates.}
Models were run at different pipeline maturity stages and, more seriously, under two different generation methods: the two-stage protocol of \S\ref{sec:pipeline} and a single-pass knowledge-grounded baseline.
Method and model are partly collinear in Table~\ref{tab:models} (three of its lowest cloze rows are models run entirely single-pass, including the one that sets the bottom of the range), and only two models were run both ways, so the aggregate spread mixes a capability effect with a method effect that we cannot cleanly separate.
Only the matched Grade-12 run is fully matched---same rules, standards, judge, and 100\% two-pass for all five models---so quantitative capability claims rest on Table~\ref{tab:g12} and not on the aggregate.
Several per-model samples are small ($n \le 250$).
The repeat-verdict audit (\S\ref{sec:reliability}) bounds the judge's random error but not systematic bias shared across repeats.
\textbf{Component attribution.}
The pipeline was iterated as a whole; we built a runnable ablation harness that isolates the knowledge dictionary, negative exemplars, and Stage-2 review, and separates open-set task structure from prompt-rule count (a minimal-rule cloze arm vs.\ a rule-inflated MCQ arm) on a fixed Grade-12 slice.
\textbf{Threshold sensitivity.}
``Pass'' binarizes a continuous score at the platform's deployment threshold (0.85); mean scores (reported throughout) show the same ordering.
\textbf{Scope.}
Results cover one subject (ELA), one curriculum (CCSS), and English; the round-by-round audit (Figure~\ref{fig:pareto}) covers one grade and generator.
\textbf{Contamination.}
We cannot rule out that generators saw CCSS-aligned item corpora in pretraining, though this would aid all formats and cannot explain the format asymmetry.
\textbf{Relation to a companion manuscript.}
A companion paper by the same authors \citep{chen2026judging} isolates the same generate-versus-verify gap on constructions with complete ground truth---mechanical string, numeric and code predicates---and measures its downstream effect on reinforcement learning with verifiable rewards.
The two manuscripts share one dataset: the corpus analysed here also serves that paper as a field signature, so its 43{,}227 scored items over 755 standards and ten generators recur there, as do two of the format pass rates of Table~\ref{tab:production} (cloze 96.48\%, MCQ 99.13\%).
The analyses of that dataset do not overlap. The pipeline of \S\ref{sec:pipeline}, the production run's headline rate, the independent-judge cross-validation of \S\ref{sec:reliability} and the distillation probe of \S\ref{sec:distill} are reported here and not there; the omission-dominant key-error split, the evaluator's self-consistency on failures, and a stratified confound analysis of the same corpus are reported there and not here.
Beyond the dataset the two share only the conceptual framing of \S\ref{sec:analysis} and part of the related-work discussion. No experiment, model run, table or figure is common to both, and neither paper's claims depend on the other's results.
We flag the shared dataset so reviewers can judge the overlap directly.

\section{Conclusion}
\label{sec:conclusion}
Closing the loop between an LLM generator and a structured expert evaluator---adversarial self-review, contrastive exemplars, and mined per-standard knowledge rules---raises standards-aligned item generation to 97.8\% evaluator-pass at full-curriculum scale. That figure is throughput past one gate rather than a measured defect rate: independent judges reproduce the format ordering but agree with the deployed evaluator only weakly on the accept/reject binary, and no student-response evidence is reported here.
The same 43{,}227-item corpus shows where the paradigm ends: answer-key construction for open-ended formats is an open-set boundary-determination task on which prompt optimization plateaus, trading answer-key over-inclusion against omission rather than reducing either, while greater model capability---or an architectural division of labour between generation and discrimination---does move the ceiling.
We offer the pipeline as infrastructure, the corpus as evidence, and the asymmetry as a target: the next capability worth engineering into generation systems is not better fluency, but calibrated boundaries.

\section*{Ethics Statement}
All generated content is synthetic assessment material for K--12 ELA; no student data was used or collected.
Prompts enforce cultural neutrality and accessibility constraints, and the evaluator screens for bias, but LLM-generated items may still encode subtle biases; we recommend human review before classroom deployment and flag that pass rates here measure evaluator-defined quality, not learning outcomes.
Scaled item generation may displace some assessment-authoring work; in the deployment studied here it re-tasks experts from drafting to auditing.

\bibliographystyle{plainnat}
\bibliography{refs}

\appendix

\section{Production Results by Grade}
\label{app:grades}
\begin{table}[h]
\centering\small
\setlength{\tabcolsep}{5pt}
\begin{tabular}{lrrr}
\toprule
Grade & MCQ & MSQ & Cloze\\
\midrule
1  & 99.1 & 96.3 & 95.8\\
2  & 98.4 & 97.9 & 96.8\\
3  & 99.6 & 98.7 & 95.9\\
4  & 99.3 & 99.3 & 96.8\\
5  & 99.1 & 96.4 & 97.7\\
6  & 99.5 & 99.0 & 97.1\\
7  & 99.0 & 99.5 & 96.5\\
8  & 99.0 & 97.6 & 98.0\\
9  & 99.5 & 99.5 & 95.3\\
10 & 98.0 & 99.5 & 95.8\\
11 & 99.5 & 97.4 & 97.4\\
12 & 99.5 & 99.0 & 96.7\\
\bottomrule
\end{tabular}
\caption{Final production run: pass rate (\%) by grade and type (Claude Opus 4.6).}
\label{tab:pergrade}
\end{table}

\section{Cloze Rule Families (Excerpt)}
\label{app:rules}
The full rule set comprises a global core plus grade-band dispatch. Representative rules, abridged:
\begin{itemize}[leftmargin=1.2em,itemsep=1pt,topsep=2pt]
\item \textbf{Field sync}: every variant mentioned in the explanation must appear in the answer array; morphology (number, case) must match the stated instruction.
\item \textbf{Stem--answer firewall}: no accepted answer may appear as a substring anywhere student-visible before the blank; synonyms of the answer in the frame must be paraphrased into abstract descriptions (``hostile pruning'').
\item \textbf{Word-bank lock}: if instructions restrict answers to a displayed bank, the key array must be a subset of the bank---the single most frequent fatal error class before R4.
\item \textbf{Collocational-cue ban} (medium/hard): if the 2--3 tokens before the blank complete the phrase for an average speaker without domain knowledge, restructure.
\item \textbf{Verbatim-retrieval ban} (medium/hard): if the exact answer string is visible anywhere in the stimulus, the item is retrieval-level and must be redesigned or downgraded to easy.
\item \textbf{Cognitive step count}: easy $=1$ operation, medium $=2$, hard $\geq 3$; recount before assigning the difficulty label.
\item \textbf{Semantic substitution test}: re-insert every candidate key into the blank; delete any that shifts meaning, connotation, register, or part of speech.
\end{itemize}

\section{Stage-2 Attack Prompt (Cloze, Abridged)}
\label{app:attacks}
\begin{quote}\small
\emph{You are now a HOSTILE ADVERSARIAL REVIEWER. Your mission is to FIND FLAWS.}
\textbf{Attack 1 --- Answer-array precision}: substitute each term into the blank; remove any that fails meaning, part of speech, or register.
\textbf{Attack 2 --- Completeness}: name 3--5 terms an expert might write; add any that fit grammatically and semantically.
\textbf{Attack 3 --- Collocational cue}: could an average person complete the 2--3 words before the blank without domain knowledge? If yes, restructure.
\textbf{Attack 4 --- Self-consistency}: does the passage already answer the question, contradict the explanation, or reveal the key?
\textbf{Attack 5 --- Explanation audit}: verify letter counts by spelling them out; weaken unverified uniqueness and modality claims. Output corrected JSON only.
\end{quote}

\section{Example Mined Knowledge Rule}
\label{app:rule}
\begin{quote}\small
\textbf{Key}: \texttt{CCSS.ELA-LITERACY.L.1.1.A | fill-in}\quad
\textbf{Category}: \textsc{difficulty\_mismatch}\quad
\textbf{Severity}: 0.74\\
\textbf{Problem}: explanation claims uppercase `B' has ``one round part'' and misstates the bowl position of lowercase `p'.\\
\textbf{Correction}: uppercase `B' has two bowls attached to a vertical stem; for `p' the bowl sits above the baseline while the stem descends below it.
\end{quote}

\section{Error-Category Taxonomy}
\label{app:taxonomy}
\begin{figure}[htbp]
\centering
\includegraphics[width=0.62\textwidth]{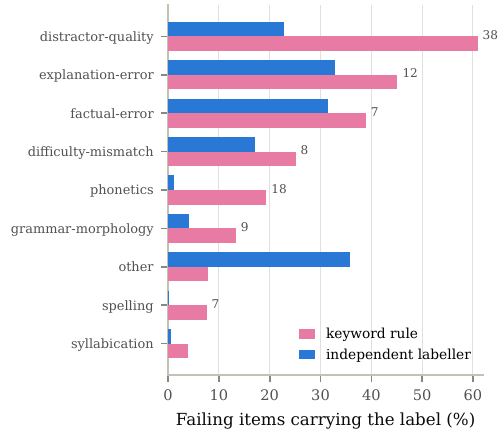}
\caption{Share of failing items carrying each error label under the keyword rule and under independent re-labelling, pooled over the three formats ($n{=}450$). Annotations give the gap in points where the rule fires more often. \textsc{distractor-quality} and \textsc{phonetics} are the largest over-firings; \textsc{other} runs the other way, the independent labeller finding no fitting class four times as often as the rule does.}
\label{fig:taxonomy}
\end{figure}

Categories are assigned by keyword matching over the judge's reasoning and suggested improvements: \textsc{phonetics} (phoneme, vowel, schwa, rhyme, \dots), \textsc{syllabication}, \textsc{dictionary\-skills} (guide word, alphabetization), \textsc{spelling}, \textsc{factual\-error}, \textsc{explanation\-error}, \textsc{distractor\-quality} (distractor, plausib-), \textsc{grammar\-morphology}, \textsc{etymology}, \textsc{difficulty\-mismatch} (rigor, too easy/hard); unmatched feedback maps to \textsc{other}. An item may receive multiple labels; Table~\ref{tab:errors} reports label shares.
Reliability of this instrument was measured by re-labelling 450 failing items (150 per format, stratified because failing items are predominantly cloze) with GPT-5.1 from the same feedback text under the same definitions.
Table~\ref{tab:taxrel} gives the per-class result and Figure~\ref{fig:taxonomy} the same data as label shares.
Two patterns are visible.
The rule's recall is high wherever it fires at all (0.53--1.00) while its precision is often very low, which is the signature of over-firing rather than under-coverage: a substring rule cannot tell an assertion from a mention, and evaluator feedback on an ELA item routinely names phonetic or orthographic properties while faulting something else.
Conversely the independent labeller assigns \textsc{other} far more often than the rule does (36\% vs.\ 8\% pooled across formats), so the ten classes leave a substantial part of the failure space undescribed.
Table~\ref{tab:taxprofile} carries this through to the per-format profiles, which is what the paper's comparative claims rest on: the cloze difficulty-mismatch concentration survives the check and the MCQ distractor-quality concentration does not.

\begin{table}[htbp]
\centering\small
\setlength{\tabcolsep}{4pt}
\begin{tabular}{lrrrrr}
\toprule
Category & Rule fires & Independent & Precision & Recall & F1\\
\midrule
\textsc{factual-error} & 175 & 142 & 0.76 & 0.94 & 0.84\\
\textsc{explanation-error} & 203 & 148 & 0.69 & 0.95 & 0.80\\
\textsc{other} & 35 & 161 & 0.51 & 0.11 & 0.18\\
\textsc{difficulty-mismatch} & 113 & 77 & 0.37 & 0.55 & 0.44\\
\textsc{distractor-quality} & 274 & 103 & 0.23 & 0.60 & 0.33\\
\textsc{syllabication} & 18 & 3 & 0.17 & 1.00 & 0.29\\
\textsc{grammar-morphology} & 60 & 19 & 0.17 & 0.53 & 0.25\\
\textsc{phonetics} & 87 & 5 & 0.06 & 1.00 & 0.11\\
\textsc{spelling} & 34 & 1 & 0.03 & 1.00 & 0.06\\
\textsc{dictionary-skills} & 16 & 0 & 0.00 & --- & ---\\
\textsc{etymology} & 18 & 0 & 0.00 & --- & ---\\
\bottomrule
\end{tabular}
\caption{Reliability of the keyword error taxonomy, measured by re-labelling 450 failing items (150 per format) with an independent model from the same feedback text
under the same category definitions. ``Rule fires'' and ``Independent'' count items receiving the
label under each labeller. Precision is the share of the rule's firings the independent labeller
confirms. High recall alongside low precision is the signature of over-firing: the rule triggers
on any mention of a category's vocabulary. Overall primary-label $\kappa$ = 0.27, mean Jaccard 0.37, exact label-set agreement 0.14.}
\label{tab:taxrel}
\end{table}

\begin{table}[htbp]
\centering\small
\setlength{\tabcolsep}{3.5pt}
\begin{tabular}{lrrrrrr}
\toprule
& \multicolumn{2}{c}{MCQ} & \multicolumn{2}{c}{MSQ} & \multicolumn{2}{c}{Cloze}\\
\cmidrule(lr){2-3}\cmidrule(lr){4-5}\cmidrule(lr){6-7}
Category & Rule & Indep. & Rule & Indep. & Rule & Indep.\\
\midrule
\textsc{distractor-quality} & 90.7 & 22.0 & 82.0 & 17.3 & 10.0 & 29.3\\
\textsc{difficulty-mismatch} & 17.3 & 12.7 & 22.0 & 2.0 & 36.0 & 36.7\\
\textsc{explanation-error} & 52.7 & 40.7 & 50.0 & 42.0 & 32.7 & 16.0\\
\textsc{factual-error} & 50.0 & 40.0 & 44.7 & 44.7 & 22.0 & 10.0\\
\textsc{phonetics} & 22.0 & 2.0 & 18.7 & 0.7 & 17.3 & 0.7\\
\textsc{syllabication} & 4.7 & 0.7 & 7.3 & 1.3 & 0.0 & 0.0\\
\textsc{spelling} & 12.7 & 0.7 & 6.7 & 0.0 & 3.3 & 0.0\\
\textsc{grammar-morphology} & 13.3 & 6.0 & 10.0 & 1.3 & 16.7 & 5.3\\
\textsc{other} & 0.7 & 21.3 & 1.3 & 43.3 & 21.3 & 42.7\\
\bottomrule
\end{tabular}
\caption{Share of failing items carrying each label (\%), by format, under the keyword rule and
under independent re-labelling ($n{=}150$ per format). The paper draws two comparative claims
from this profile. Difficulty mismatch concentrating in cloze survives (+18.7 points by rule, +24.0 independently). Distractor quality concentrating in MCQ does not (+80.7 by rule but -7.3 independently, a reversal), because evaluator feedback on an MCQ discusses the
distractors whatever the defect. Shares are per item and do not sum to 100, since an item may
carry several labels.}
\label{tab:taxprofile}
\end{table}

\section{Corpus Composition}
\label{app:composition}
Table~\ref{tab:composition} gives the shape of the 9{,}074-item production run: items per grade and format on the left, pass rate by format and labelled difficulty on the right.
Cells are near-balanced by construction, with grades 1, 3, 9, and 12 carrying extra cloze items from re-runs inside the production window.
The difficulty columns deserve a caution that Table~\ref{tab:difficulty} also carries: ``easy'' and ``hard'' are the requested labels sent to the generator, and the near-identical pass rates across them say that the evaluator judged difficulty-appropriateness to be met about equally often in each condition. They say nothing about whether an item labelled hard is empirically harder for a student, which requires response data we do not have (\S\ref{sec:limitations}).

\begin{table}[htbp]
\centering\small
\setlength{\tabcolsep}{5pt}
\begin{tabular}{lrrrr|lrr}
\toprule
Grade & MCQ & MSQ & Cloze & Total & Type\,$\times$\,difficulty & $n$ & Pass\%\\
\midrule
1 & 216 & 216 & 432 & 864 & Cloze, easy & 1,267 & 96.53\\
2 & 189 & 189 & 188 & 566 & Cloze, hard & 1,250 & 95.44\\
3 & 237 & 237 & 584 & 1058 & Cloze, medium & 1,261 & 97.46\\
4 & 408 & 408 & 406 & 1222 & MCQ, easy & 880 & 99.43\\
5 & 219 & 219 & 218 & 656 & MCQ, hard & 882 & 99.09\\
6 & 207 & 207 & 207 & 621 & MCQ, medium & 882 & 98.87\\
7 & 198 & 198 & 198 & 594 & MSQ, easy & 884 & 98.19\\
8 & 204 & 204 & 204 & 612 & MSQ, hard & 884 & 98.30\\
9 & 195 & 195 & 382 & 772 & MSQ, medium & 884 & 98.64\\
10 & 195 & 195 & 191 & 581 & & &\\
11 & 192 & 192 & 192 & 576 & & &\\
12 & 184 & 192 & 576 & 952 & & &\\
\bottomrule
\end{tabular}
\caption{Composition of the 9{,}074-item production run. Left: items per grade and format.
Right: pass rate by format and labelled difficulty. Grades 1, 3, 9, and 12 carry extra cloze
items from re-runs during that window. Difficulty is the requested label, not a measured
statistic; that easy and hard items pass at indistinguishable rates is a statement about the
evaluator's judgment, not about empirical item difficulty (\S\ref{sec:limitations}).}
\label{tab:composition}
\end{table}

\section{Generation-Method Audit}
\label{app:method}
The corpus records a \texttt{generation\_method} field on every item, and it takes two values: \texttt{two\_pass\_verify}, the architect-then-adversary protocol this paper proposes, and \texttt{single\_pass\_knowledge\_grounded}, an earlier single-pass baseline that retrieves knowledge rules but runs no adversarial second stage.
The aggregate model table (Table~\ref{tab:models}) pools both, which materially weakens it as a model comparison.
Table~\ref{tab:method} splits cloze pass rate by model and method.

Three models were run exclusively single-pass, including the one at the bottom of the aggregate range, and three exclusively two-pass at the top, so method and model are partly collinear and the aggregate spread cannot be attributed to capability alone.
Only two models were measured both ways, and they disagree in sign: Kimi K2.5 gains 6.2 points under the two-pass protocol ($n{=}429$ vs.\ $4{,}665$) while GPT-OSS-120B loses 3.7 ($n{=}169$ vs.\ $85$).
Two models with opposing signs and unbalanced samples do not support an estimate of the method effect; they support only the negative conclusion that the effect is not obviously negligible.
This is why the paper's quantitative capability claims are stated from the matched comparison of Table~\ref{tab:g12}, where the rule set, standards, judge, and grade are fixed and all five models are 100\% two-pass, and why the aggregate table is presented as observational.

\begin{table}[htbp]
\centering\small
\setlength{\tabcolsep}{4pt}
\begin{tabular}{lrlrlr}
\toprule
& \multicolumn{2}{c}{Two-pass (this paper)} & \multicolumn{2}{c}{Single-pass baseline} & \\
\cmidrule(lr){2-3}\cmidrule(lr){4-5}
Model & $n$ & Pass\% [95\% CI] & $n$ & Pass\% [95\% CI] & $\Delta$\\
\midrule
Claude Opus 4.6 & 3,969 & 96.3 [95.7, 96.8] & --- & --- & ---\\
Claude Opus 4.5 & 191 & 93.2 [88.7, 96.0] & --- & --- & ---\\
Gemini 3.1 Pro & 6,897 & 87.8 [87.0, 88.5] & --- & --- & ---\\
Gemini 3 Flash & 83 & 86.8 [77.8, 92.4] & --- & --- & ---\\
GLM-5 & --- & --- & 408 & 86.5 [82.9, 89.5] & ---\\
Kimi K2.5 & 429 & 84.2 [80.4, 87.3] & 4,665 & 77.9 [76.7, 79.1] & $+6.2$\\
GPT-5.4 & 192 & 82.8 [76.8, 87.5] & --- & --- & ---\\
Unidentified generator & --- & --- & 288 & 78.5 [73.4, 82.8] & ---\\
GPT-OSS-120B & 169 & 72.8 [65.6, 78.9] & 85 & 76.5 [66.4, 84.2] & $-3.7$\\
Gemini 3.1 Flash-Lite & 623 & 65.0 [61.2, 68.7] & --- & --- & ---\\
MiniMax M2.5 & --- & --- & 104 & 59.6 [50.0, 68.5] & ---\\
\bottomrule
\end{tabular}
\caption{Cloze pass rate split by the generation method recorded for each item, the audit behind
the caveat on Table~\ref{tab:models}. Wilson intervals. Only two models were run both ways, with
opposite signs, so the method effect cannot be estimated reliably; what the table does show is
that method and model are partly collinear, since several models were run exclusively one way.
The matched comparison of Table~\ref{tab:g12} is 100\% two-pass for all five of its models.}
\label{tab:method}
\end{table}

\section{Independent-Judge Cross-Validation: Details}
\label{app:indep}
This appendix supports \S\ref{sec:indep}.
The sample is 390 items: an \emph{asymmetry} stratum of 100 items per format drawn at random from the matched production run, and an \emph{agreement} stratum of 30 deployed failures per format.
The second stratum exists because the deployed failure rate is 2--4\%, so a proportional sample of a few hundred items would contain too few rejections for $\kappa$ to be estimable at all; the cost is that raw agreement statistics on this sample describe a corpus far more defective than the real one, which is why Table~\ref{tab:indepdetail} reports a prevalence-corrected $\kappa$ alongside the raw value.
Each judge saw the item stem, options where present, answer key, provided explanation, target standard with its description, and grade, and saw neither the deployed verdict nor the deployed score nor which stratum the item came from. Item order was a single fixed shuffle shared by all three judges. Temperature was zero.

The rubric asked for a single accept/reject decision plus one defect code from \{key wrong, key incomplete, ambiguous, off-standard, off-grade, other\}, with the instruction to reject only on defects serious enough to block deployment and explicitly not to penalize phrasing the judge would have chosen differently.
The rubric is our own rather than the deployed evaluator's criteria by design: supplying the target judge's rubric would manufacture agreement, and the question is whether the format effect survives a different notion of quality, not whether one rubric can be replicated.

\begin{figure}[htbp]
\centering
\includegraphics[width=0.62\textwidth]{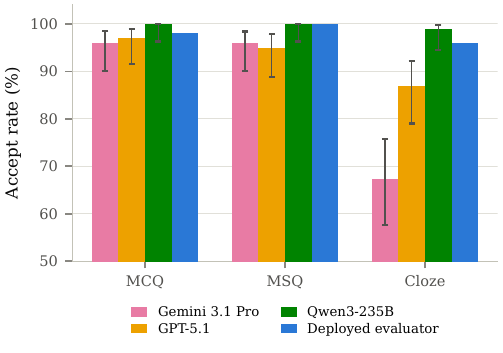}
\caption{Accept rate by item type for each independent judge and for the deployed evaluator, on the asymmetry stratum ($n{=}100$ per type, Wilson 95\% intervals). MCQ and MSQ sit near ceiling for every judge; cloze separates them. The deployed evaluator is the most permissive of the four on cloze, so the format gap this paper reports is the smallest of the four measurements of it.}
\label{fig:indepjudges}
\end{figure}

Figure~\ref{fig:indepjudges} gives the accept rates: the format ordering is the same under every judge, and the two frontier judges recover a larger cloze deficit than the deployed evaluator.
Table~\ref{tab:indepdetail} gives the agreement statistics together with agreement among the judges themselves, which is the necessary control. Gemini 3.1 Pro and GPT-5.1 agree with each other at $\kappa{=}0.40$ and with the deployed evaluator at 0.43 and 0.33; pairs involving Qwen3-235B sit near 0.11. Low judge-versus-deployed agreement is therefore not evidence against the deployed evaluator specifically: binary item acceptability is a low-agreement judgment among competent judges, which bounds what any single-judge pass rate can certify.

Table~\ref{tab:indepdefects} breaks the rejections down by cited defect, restricted to items the deployed evaluator passed.
Answer-key incompleteness is the dominant charge on cloze for both frontier judges and is rare on MCQ, which is the specific failure the open-set analysis of \S\ref{sec:analysis} predicts should survive a pipeline optimized against a single-answer notion of correctness.
Requiring both frontier judges to concur, 9.9\% of cloze items that passed the deployed evaluator are rejected, against 1.0\% of MCQ and none of MSQ.

Qwen3-235B deserves separate comment because it is the one judge that fails.
It accepts 99.7\% of everything it sees and recovers 8.3\% of deployed failures, so as an evaluator it reproduces the verdict form while carrying almost no accept/reject boundary.
That is the dissociation the distillation study of \S\ref{sec:distill} induces deliberately in a 14B student, here in a larger off-the-shelf model: the generative surface of judging is easy to reproduce and the discriminative boundary is not.
It also serves as a check on the design, since a judge that accepts everything cannot manufacture the format effect the other two report.

\begin{table}[htbp]
\centering\small
\setlength{\tabcolsep}{4pt}
\begin{tabular}{lrrrrr}
\toprule
Judge & Agree\% & $\kappa$ raw & $\kappa$ prev.-corr. & Fail-recall & Unusable\\
\midrule
google/gemini-3.1-pro-preview & 79.2 & 0.428 & 0.126 & 55.9\% & 1.5\%\\
openai/gpt-5.1 & 78.7 & 0.330 & 0.140 & 35.4\% & 0.0\%\\
qwen/qwen3-235b-a22b & 77.2 & 0.115 & 0.133 & 8.3\% & 0.0\%\\
\midrule
\emph{gemini-3.1-pro-preview vs.\ gpt-5.1} & 81.6 & 0.404 & --- & --- & ---\\
\emph{gpt-5.1 vs.\ qwen3-235b-a22b} & 86.3 & 0.124 & --- & --- & ---\\
\emph{gemini-3.1-pro-preview vs.\ qwen3-235b-a22b} & 77.7 & 0.103 & --- & --- & ---\\
\bottomrule
\end{tabular}
\caption{Agreement of each independent judge with the deployed evaluator's accept/reject verdict
over all 390 sampled items, and, below the rule, agreement of the judges with each other.
$\kappa$ raw is computed on the fail-oversampled sample; the prevalence-corrected column
reweights each item type to its production failure rate, which is the figure that describes the
deployed corpus. ``Fail-recall'' is the share of deployed failures the judge also rejects.
``Unusable'' is the share of replies that were empty or unparseable, reported because a high rate
there would make every other column meaningless.}
\label{tab:indepdetail}
\end{table}

\begin{table}[htbp]
\centering\small
\setlength{\tabcolsep}{3.5pt}
\begin{tabular}{llrrrrrr}
\toprule
Judge & Type & Key inc. & Key wrong & Ambig. & Off-std & Off-grade & Other\\
\midrule
gemini-3.1-pro-preview & MCQ & 0 & 1 & 2 & 1 & 0 & 0\\
gemini-3.1-pro-preview & MSQ & 0 & 0 & 0 & 0 & 0 & 4\\
gemini-3.1-pro-preview & Cloze & 18 & 0 & 1 & 6 & 0 & 6\\
gpt-5.1 & MCQ & 2 & 0 & 0 & 1 & 0 & 0\\
gpt-5.1 & MSQ & 4 & 1 & 0 & 0 & 0 & 0\\
gpt-5.1 & Cloze & 12 & 0 & 0 & 0 & 1 & 0\\
qwen3-235b-a22b & Cloze & 0 & 1 & 0 & 0 & 0 & 0\\
\bottomrule
\end{tabular}
\caption{Defects the independent judges cite when rejecting an item that the deployed evaluator
\emph{passed} (asymmetry stratum, $n{=}100$ per type). Answer-key incompleteness dominates the
cloze column for both frontier judges and is rare on MCQ, which is the failure mode the open-set
analysis of \S\ref{sec:analysis} predicts. Qwen3-235B rejects almost nothing and so contributes
almost no rows.}
\label{tab:indepdefects}
\end{table}

\end{document}